\documentclass{article}
\usepackage{spconf,amsmath,amssymb,graphicx}
\usepackage{color}
\usepackage{cite}
\usepackage{balance}
\usepackage{wasysym}
\usepackage{bm}
\usepackage{booktabs}
\usepackage{etoolbox,siunitx}
\robustify\bfseries
\usepackage{caption}
\usepackage{enumitem} 
\usepackage{listings}
\usepackage{pifont}
\usepackage[T1]{fontenc}
\usepackage{algpseudocode}
\usepackage{algorithm, setspace}
\makeatletter
\algnewcommand{\LineComment}[1]{\Statex \hskip\ALG@thistlm \(\triangleright\) #1}
\algnewcommand{\IndentLineComment}[1]{\Statex \hskip\ALG@tlm \(\triangleright\) #1}

\def\LL{{\mathcal{L}}}

\def\UU{{\mathcal{U}}}
\def\PP{{\mathcal{P}}}

\def\EE{{\mathcal{E}}}
\def\AA{{\mathcal{A}}}

\def\WW{$\tt wav2vec2$}
\def\WB{{$\tt w2v\_base$}}
\def\WR{{$\tt w2v\_rob$}}
\def\WL{{$\tt w2v\_large$}}
\def\WLS{{$\tt w2v\_large\_sup$}}
\def\XLSR{{$\tt xlsr$}}
\def\TLL{{$\tt PT \rightarrow FT$}}
\def\PT{$\tt PT$}
\def\FT{{$\tt FT$}}
\def\MPT{{$\tt mPT$}}

\def\DUST{{$\tt DUST$}}
\def\TL{{$\tt TL$}}

\DeclareFontFamily{U}{matha}{\hyphenchar\font45}
\DeclareFontShape{U}{matha}{m}{n}{
      <5> <6> <7> <8> <9> <10> gen * matha
      <10.95> matha10 <12> <14.4> <17.28> <20.74> <24.88> matha12
      }{}
\DeclareSymbolFont{matha}{U}{matha}{m}{n}

\DeclareMathSymbol{\Lt}{3}{matha}{"CE}
\DeclareMathSymbol{\Gt}{3}{matha}{"CF}
\title{Magic dust for cross-lingual adaptation of monolingual wav2vec-2.0}
\name{Sameer Khurana$^1$, Antoine Laurent$^2$, James Glass$^1$ \thanks{This work uses HPC resources of IDRIS under the allocation AD011012527 made by GENCI.}}
\address{$^1$MIT Computer Science and Artificial Intelligence Laboratory, Cambridge, MA, USA\\ $^2$LIUM - Le Mans University, France}

\begin{document}
\ninept
\maketitle
\begin{abstract}
We propose a simple and effective cross-lingual transfer learning method to adapt monolingual wav2vec-2.0 models for Automatic Speech Recognition (ASR) in resource-scarce languages. We show that a monolingual wav2vec-2.0 is a good few-shot ASR learner in several languages. We improve its performance further via several iterations of Dropout Uncertainty-Driven Self-Training (DUST) by using a moderate-sized unlabeled speech dataset in the target language. A key finding of this work is that the adapted monolingual wav2vec-2.0 achieves similar performance as the topline multilingual XLSR model, which is trained on fifty-three languages, on the target language ASR task.
\end{abstract}
%
\section{Introduction} 
\label{sec:intro}
Few-shot learning, the ability to train a machine to exhibit intelligent behavior via a small amount of supervision has been a long-standing research goal in Artificial Intelligence. To build few-shot learners we turn to a class of transfer learning (\TL{}) methods that extract knowledge from vast quantities of unlabeled data to make the task of learning from a few labeled examples easier.
Recently, Self-Supervised Learning (SSL) has emerged as a promising \TL{} approach of learning from unlabeled data \cite{chen2020simple, devlin2019bert,  oord2019representation}.


SSL~\cite{10.5555/2987189.2987204, Schmidhuber90makingthe} refers to the process of Pre-Training (\PT{}) a model on unlabeled data using an SSL task, such as masked self-prediction \cite{devlin2019bert}. The Pre-Trained model is then Fine-Tuned (\FT{}) on the target task via a few labeled examples. Hence, SSL forms the first stage of the \PT{} then \FT{} (\TLL{}) sequential \TL{} framework \cite{wang2015transfer}. Recently, speech neural net encoders Pre-Trained using the \WW{} SSL framework have proven to be excellent few-shot learners for automatic speech recognition (ASR) across multiple languages \cite{Baevski2020wav2vec, conneau2020unsupervised}. However, \WW{} assumes access to massive amounts of unlabeled data for PT, which diminishes their usefulness to resource-scarce languages, where the \textit{massive unlabeled data} assumption does not hold.

To remedy the above issue, \cite{conneau2020unsupervised} proposes \XLSR{}, a cross-lingual sequential \TL{} framework of the form \MPT{} $\rightarrow$ \FT{}, i.e., Multilingual Pre-Training of \WW{} followed by target language ASR fine-tuning on a few labeled examples. Indeed, Pre-Trained \XLSR{} is an excellent few-shot learner for ASR in multiple languages. However, in this work we show that \XLSR{}'s ASR performance is quite poor if there is a domain mismatch between the target language speech and the speech data used to Pre-Train \XLSR{}. Thus, to make \XLSR{} a truly universal speech model, we would have to Pre-Train on speech from all languages in all possible speech domains, which is clearly an unscalable strategy. Instead, in this work, we propose a \TL{} framework that could efficiently adapt any Pre-Trained \WW{} model, monolingual or multilingual, to make it a good few-shot ASR learner in any target language in any speech domain.

In this work, motivated by the SSL framework's limitations when developing ASR for a resource-scarce language, we propose a simple yet effective cross-lingual \TL{} framework (\S \ref{sec:transfer_algo}) for \WW{} model adaptation to a target language. Our adaptation framework is a sequential \TL{} framework consisting of three steps: First, we Pre-Train a \WW{} model on a high-resource language. Second, we perform supervised fine-tuning of the Pre-Trained \WW{} model on the target language ASR task using ten hours of labeled data. Finally, we perform Dropout Uncertainty-Driven Self-Training (DUST) \cite{khurana2021unsupervised} using a hundred hours of unlabeled speech data in the target language for adaptation of the Fine-Tuned \WW{} model.

Through this work, we make the following \textbf{contributions}: 1) We analyze the cross lingual transferability of several Pre-Trained English \WW{} models (Table \ref{tab:1}) across eight target languages. We show that by simply fine-tuning English $\tt wav2vec2$ on ten hours of labeled data in target languages, we can recover on average up to 86\% of the performance of the fine-tuned multilingual \XLSR{} topline. Still, there is a considerable gap in performance between \WW{} and \XLSR{} on target languages that are considered in-domain for \XLSR{}, but the gap is much smaller on a more challenging out-of-domain Arabic target language. Another interesting finding is that ASR Fine-Tuning of the Pre-Trained \WW{} models on labeled data in the source language (English) before Fine-Tuning on the target languages hurts cross-lingual transfer. 2) We adapt an English \WW{} model to two target languages, French and Arabic, under the constraint that in each target language we have ten hours of labeled data for ASR training and a hundred hours of unlabeled data for adaptation. For French, we show that by starting with a Pre-Trained English \WW{} model and applying the proposed adaptation procedure (\S \ref{sec:transfer_algo}), we are able to reach similar ASR performance as the \XLSR{} topline. For Arabic, both the \XLSR{} and English \WW{} perform poorly and hence, we apply the adaptation procedure to both the models and improve the ASR performance considerably. A key finding of this study is that it is possible to adapt a monolingual \WW{} model Pre-Trained on a high-resource language using moderately-sized unlabeled data and small-sized labeled data in the target language to achieve similar performance as the multilingual \WW{} model Pre-Trained on multiple languages. Although the amount of unlabeled data that we use for adaptation is orders of magnitude smaller than the data used to Pre-Train \WW{} models, a moderate-sized unlabeled dataset might not be available for extremely resource-scarce and endangered languages. This scenario is out of scope for this paper.

\section{Method}
\paragraph*{Self-Training} Self-Training (ST) \cite{scudder1965probability} is a Teacher/Student (T/S) \TL{} framework that leverages unlabeled data by pseudo-labeling it. ST proceeds by building a base model, known as teacher, using the labeled data. The teacher is used to predict (pseudo-)labels for the unlabeled data points. Then, a new model, known as student, is trained on the combined labeled and pseudo-labeled data points. Due to having access to more supervision, the student is expected to generalize better than the teacher on the task at hand. ST is an iterative process, where, the student from a previous round becomes the teacher for the next round of ST. Recently, ST has shown excellent results in neural sequence generation tasks such as ASR \cite{Kahn_2020, xu2020iterative, khurana2021unsupervised} and Machine Translation \cite{he2019revisiting}
\paragraph*{Transfer Learning Algorithm}
\label{sec:transfer_algo}
The overall transfer learning process is described in Algorithm 1. We assume access to a set $\LL_T$ of labeled examples and a set $\UU_T$ of unlabeled speech utterances in the target language. Also, we are given a set $\UU_S$ of unlabeled speech utterances in the source language. The transfer learning process proceeds by Pre-training a neural network $f_{\phi, p}$ on unlabeled source language set $\UU_S$ with dropout layers, using a dropout probability $p \in [0, 1]$. The Pre-training process leads to the initial model $f_{\phi_0, p}$, which is Fine-Tuned on the target language labeled set $\LL_T$ to give the first-generation teacher model $f_{\phi_1, p}$ for Dropout-Uncertainty driven Self-Training (\DUST{}). Next, the base teacher model $f_{\phi_1, p}$ is used to provide predictions on the target language unlabeled set $\UU_T$ to provide pseudo-parallel data of which a subset $\PP$ is chosen based on the model's uncertainty about its predictions on each unlabeled data point $x_u \in \UU_T$. Finally, a student model, is trained on the combined labeled $\LL_T$ and pseudo-labeled set $\PP$. We perform N iterations of the Teacher/Student training, where the student $f_{\phi_n, p}$ from the $n^{th}$ iteration becomes teacher for the $(n+1)^{th}$ iteration. Usually, in each iteration of \DUST{}, a randomly initialized neural network is used as the student model, but, in our adaptation framework, the Pre-Trained source language SSL model $f_{\phi_0, p}$ is used as the student in each \DUST{} iteration.
\begin{algorithm}[t]
\setstretch{1.}
\label{algo:1}
\caption{Transfer Learning Algorithm}
\begin{algorithmic}[1]
\State Given labeled data $\LL_S$ and unlabeled data $\UU_S$ in the source language
\State Given labeled data $\LL_T$ and unlabeled data $\UU_T$ in the target language
\State Given $R$ natural numbers 
\State Pre-Train $f_{(\phi, p)}$ on $\UU_S$ to get $f_{(\phi_0, p)}$
\State Fine-Tune $f_{(\phi_0, p)}$ on $\LL_T$ to get $f_{(\phi_1, p)}$
\For{n =1\ \text{to}\ N}
\State $f_{(\phi_{n+1}, p)}$ = \DUST{}($f_{(\phi_n, p)}$, $f_{(\phi_0, p)}$, $\LL_T$, $\UU_T$)
\EndFor
\Function{\DUST{}}{$g^{\text{Teacher}}_{(\theta, p)}, f^{\text{Student}}_{(\psi, p)}, \LL, \UU$}
\State Let $\PP$ be the set of selected pseudo-labeled data points
\State Let $\EE$ be a set of edit distances
\State Initialize $\PP$ and $\EE$ as empty sets
\ForAll{$x_u \in \UU$}
\State Compute deterministic forward pass $g^{\text{Teacher}}_{(\theta, 0)}(x_u)$
\State $\hat{y}^{\text{ref}}_u = \text{beam\_search}(g^{\text{Teacher}}_{(\theta, 0)}(x_u))$ 
\ForAll{$r \in R$}
\State Set random seed to $r$
\State Compute stochastic forward pass $g^{\text{Teacher}}_{(\theta, p)}(x_u)$
\State $\hat{y}^r_u = \text{beam\_search}(g^{\text{Teacher}}_{(\theta, p)}(x_u))$ 
\State $e = \text{edit\_distance}(\hat{y}_u^r, \hat{y}_u^{\text{ref}})$
\State Add $e$ to the set $\EE$
\EndFor
\If{$\text{max}(\EE) < \tau |\hat{y}_u^{\text{ref}}|$ (with $\tau$ a filtering threshold)} 
\State Add $\{(x_u, \hat{y}_u^{\text{ref}}), (x_u, \hat{y}_u^0), \ldots, (x_u, \hat{y}_u^R\})$ to $\PP$
\EndIf
\EndFor
\State Fine-Tune $f^{\text{Student}}_{(\psi, p)}$ on $\AA=\LL \cup \PP$
\State \Return $f^{\text{Student}}_{(\psi, p)}$
\EndFunction
\end{algorithmic}
\end{algorithm}

DUST performs pseudo-label filtering by measuring the model's confidence about its predictions on the unlabeled points $x_u \in \UU_T$. The filtering process for a particular unlabeled example $x_u$ consists of the following steps: 1) First, we generate a reference hypothesis $\hat{y}^{u}_{\text{ref}}$ for the unlabeled instance $x_u$ using beam search. During inference, the model's dropout layers are deactivated and hence, this step imitates the usual ASR inference process. 2) Second, we sample $R$ hypotheses $(\hat{y}^{r}_u)_{r=1}^{R}$ from the model by running beam search $R$ times with a different random seed $r \in R$ each time. During inference, the dropout layers are active, hence each beam search iteration would lead to a slightly different hypothesis. This is akin to getting predictions from different models. 3) Finally, we compute the Levenshtein edit distance \cite{1966SPhD...10..707L} normalized by the length of the reference hypothesis between each of the $R$ stochastically sampled hypothesis and the one reference hypothesis, which gives us a set $\EE$ of $R$ edit distances. If all the edit-distances in $\EE$ are less than the threshold ratio $\tau$ of the length $|\hat{y}_t^{\text{ref}}|$ of the reference hypothesis, then we add the pseudo-labeled data points $\{(x_u, \hat{y}_u^{\text{ref}}), (x_u, y^{0}_u), \ldots, (x_u, y^{R}_u\})$ to $\PP$,  otherwise we reject it. In practice, we set $R=3$ and hence, we have a total of four hypotheses per utterance. Unlike the original DUST that adds only the reference pseudo-label hypothesis for $x_u$ to the set $\PP$, we also add the sampled hypotheses. Adding multiple pseudo-labels corresponding to an unlabeled instance $x_u$ for student model training could increase model's robustness to noise in pseudo-labels. This idea is also explored in \cite{dey2019exploiting}.

\begin{table*}[t]
    \caption{Cross-Lingual Transferability of Pre-Trained \WW{} model on eight target languages. Seven languages are from the MLS dataset of read audiobooks, while Arabic is from the MGB broadcast news dataset}
    \centering
    \sisetup{table-format=2.1,round-mode=places,round-precision=1,table-number-alignment = center,detect-weight=true,detect-inline-weight=math}
    \resizebox{.99\linewidth}{!}{%
    \setlength{\tabcolsep}{5pt}
    \renewcommand{\arraystretch}{1.4}
    \begin{tabular}{llcccccccccccc}\toprule
        &&&\multicolumn{8}{c}{WER / CER [\%]}&&&WERR / CERR\\
        \cmidrule{3-12} \cmidrule{14-14}
         \textbf{Target Langs}& & MLS/en & MLS/fr & MLS/de & MLS/it & MLS/pl & MLS/es & MLS/pt & MLS/nl & MGB/ar \\
        Model&PT& &  &  &  &  &  &  &  & & Avg.$\downarrow$ & & Avg.$\uparrow$\\\midrule
        Baseline & &119.1 / 58.5& 114.2 / 51.6& 106.0 / 41.5&99.5 / 35.0& 111.9 / 44.7& 99.5 / 37.3&107.0 / 45.3& 108.8 / 50.0&112.0 / 51.5& 107.4 / 44.6&& 0 / 0 \\
        \WB{} & LS960 & 23.4 / 8.1 & 44.0 / 14.5& 28.6 / 6.9 & 34.1 / 7.3 & 35.6 / 6.9 & 37.2 / 8.6 & 41.1 / 10.9 & 47.2 / 14.2 & 47.4 / 15.1 & 39.4 / 10.6& & 79.0 / 87.8 \\
        \WL{} & LS960 & 17.1 / 5.8 & 40.9 / 13.3& 28.3 / 6.8 & 33.3 / 6.9 & 32.0 / 6.2 & 23.6 / 5.6 &  38.6 / 10.2 & 45.0 / 13.3& 42.7 / 14.2 & 35.6 / 9.6&&  83.6 / 90.5 \\
        \WL{} & LL60k &12.3 / 4.0 & 39.9 / 12.7 & 26.7 / 6.4 & 31.8 / 6.7 & 32.8 / 6.4 & 21.9 / 5.1 & 35.6 / 9.4 & 42.6 / 12.6 & 42.0 / 13.2 & 34.2 / 9.1&& 85.6 / 92.1  \\
        \WR{} & LL60k+ & 12.8 / 4.2 & 38.3 / 12.3 & 26.7 / 6.4 & 30.3 / 6.2 & 34.2 / 6.6 & 22.9 / 5.3 & 34.2 / 8.9 & 39.1 / 11.8 & 41.6 / 13.1&33.4 / 8.8&& \textbf{86.3 / 92.5 } \\
        \WLS{} & LL60k& \textbf{7.6 / 2.5} & 44.2 / 14.2 & 31.1 / 7.2 & 37.7 / 7.8 & 46.5 / 9.0 & 28.1 / 6.4 & 40.8 / 10.4 & 51.3 / 15.3 & 50.6 / 15.8 &41.3 / 10.8&& 78.8  / 88.6 \\
        Topline (\XLSR{}) & MLS+& 17.6 / 6.3 & \textbf{19.7  / 6.5} & \textbf{11.1 / 3.1} & \textbf{17.1 / 3.6} & \textbf{16.4 / 3.3} & \textbf{7.9 / 2.1} & \textbf{20.4 / 5.3} & \textbf{21.7 / 6.3} & \textbf{37.9 / 12.0}&\textbf{19.0 / 5.3}&& 100 / 100 
    \end{tabular}
    }
    \label{tab:1}
\end{table*}

\paragraph*{Pre-Training}
\label{sec:pt}
In our work, we explore the following Pre-Trained \WW{} SSL models that provide the initial model $f_{\phi_0, p}$ (Algorithm 1) for transfer learning.
\begin{itemize}
\item \textbf{Wav2Vec2.0 Base} (\WB{}) \cite{Baevski2020wav2vec}: consists of 0.1 billion parameters and is Pre-Trained on the Librispeech 960 hours (LS960) \cite{librispeech} English speech dataset in the read speech domain.
\item \textbf{Wav2Vec2.0 Large} (\WL{}) \cite{Baevski2020wav2vec}: consists of 0.3 billion parameters and is Pre-Trained on either LS960 or Libri-Light 60k (LL60k) hours \cite{librilight} English read speech dataset. 
\item \textbf{Wav2Vec2.0 Robust} (\WR{}) \cite{hsu2021robust}: consists of the same architecture as the large model but, is trained on three speech datasets namely Switchboard (SWBD), English part of CommonVoice (CV-En) and LL60k. We refer to the combination of these three datasets as LL60k+.
\item \textbf{XLSR-53} (\XLSR{}) \cite{conneau2020unsupervised}: consists of the same architecture as w2v-large which is trained on the following datasets Multilingual Speech (MLS), BABEL and CommonVoice (CV), that combined consists of 53 languages. We refer to the combination of these three datasets as MLS+.
\end{itemize}
We use the publicly available Pre-Trained \WW{} model checkpoints from fairseq toolkit \cite{ott2019fairseq}.
\paragraph*{Fine-Tuning}
\label{sec:ft}
The Fine-Tuning of Pre-Trained SSL models consists of 1) Adding a linear projection layer $h_{\alpha}: \mathbb{R}^{T \times d} \rightarrow \mathbb{R}^{T \times |V|}$ to the output of the SSL model, where $V$ is the output character vocabulary for the task of ASR, 2) ASR task Fine-Tuning of only the projection layer for the first $k$ training iterations and 3) Joint ASR task Fine-Tuning of both the SSL model and the projection layer until convergence. Note the \WW{} SSL models consists of a Convolutional Neural Network (CNN) feature extractor, followed by a transformer encoder. The CNN feature extractor remains frozen throughout the ASR Fine-Tuning process.
\section{Experiment Setup}
\paragraph*{Transfer Learning Targets}
We chose seven languages from the MLS dataset as the targets for cross-lingual adaptation of the Pre-Trained \WW{} SSL models, namely French (MLS/fr), German (MLS/de), Italian (MLS/it), Polish (MLS/pl), Spanish (MLS/es), Portugese (MLS/pt) and Dutch (MLS/nl). In addition, we also target Arabic from the Multi-Genre Multi-Dialectal Broadcast News (MGB) dataset \cite{ali2019mgb2}. In order to simulate the resource-scarce ASR scenario, we assume access to just ten hours of labeled data and a hundred hours of unlabeled data in each target language. We use the official nine hours labeled split in MLS for training and the one hour split for validation. We report Word Error Rates (WERs) on the unseen development set. The hundred hours unlabeled set is sampled randomly from the full training set (minus the utterances in the ten hours split). For Arabic, we randomly sample ten hours of labeled data, of which nine hours is used for training and one hour for validation. We also randomly sample a hundred hours of speech from the 1200 hours MGB training set for cross-lingual adaptation. The results are reported on the standard development set. For the \XLSR{} model, MGB/ar is considered an out-of-domain target language because \XLSR{} is Pre-Trained on multiple datasets including MLS, which are in the read speech and conversational domains, while MGB is in the broadcast news domain. This is evident from the high WERs of the Fine-Tuned \XLSR{} on the MGB/ar dataset as compared to the MLS target languages in Table \ref{tab:1}.  
\paragraph*{Hyperparameters For ASR Fine-Tuning}
ASR Fine-Tuning of the Pre-Trained SSL models is performed on the ten hours labeled data $(x,y) \in \LL_T$ in the target language $T$, where $x$ is the input speech waveform and $y$ is the corresponding sub-word token sequence. We choose characters as sub-word units for ASR training. The model is trained using the Connectionist Temporal Classification (CTC) \cite{Graves12} loss. For optimization, we use the Adam optimizer with a learning rate schedule given by the following equation:
\begin{align*}
\text{lr} = \text{max\_lr} * \text{warmup\_steps}^{0.5} * \\ \text{min}(\text{step}^{-0.5}, \text{step} * \text{warmup\_steps}^{-1.5})
\end{align*}
where, max\_lr is the maximum learning rate, warmup\_steps are the number of training iterations before the maximum learning rate is achieved and step refers to the current training iteration. We use a relatively small value of 1e-4 for max\_lr and the first 8k training iterations for warmup. The model is trained for a total of 300 epochs. For the first 4k training iterations, we only train the linear projection layer $h_\alpha$. Batching is performed by pooling together raw speech waveforms in such a way that the total number of samples do not exceed 3.2 million. We use a gradient accumulation factor of four to ensure that the model is updated after every four training iterations, which leads to an effective batch size that is four times the original. The feature sequence output by the CNN encoder of the SSL models is randomly masked in the time dimension. We mask a span of ten consecutive time steps with a masking probability of 0.65, which leads to 65\% of the input signal being masked. We use 4 V100-32GB GPUs for fine-tuning. We use the Espnet2 codebase \cite{watanabe20202020} to perform all our experiments.
\paragraph*{Decoding}
We use beam search decoding without a language model (LM) with a beam size of 10. We do not use an LM because, in this work we are solely concerned about the acoustic model adaptation. Also, in a resource-scarce ASR scenario, we might not have text data to train a LM.

\section{Results}
\begin{table}[t]
    \centering
    \caption{Transfer of Pre-Trained \WR{} to the target French language in the MLS dataset}
    \resizebox{.9\linewidth}{!}{%
    \begin{tabular}{lccccc} \toprule
          &  & \multicolumn{2}{c}{WER [\%]} & & WERR [\%] \\
          \cmidrule{3-4} \cmidrule{6-6}
         Method & $|\PP|$ [k] & $\PP$ & MLS / fr & &MLS / fr \\ \midrule
         Baseline (\WR{}) & & & 38.3 && 0\\
         DUST1 & 11 & 20.2& 31.9 & &34.4\\
         DUST2 & 24& 20.3& 27.4 & &58.6\\
         DUST3 & 30 & 20.0& 24.2 & &75.8\\
         DUST4 & 30& 19.2& 23.5 & &79.6\\
         \textbf{DUST5} & 30& \textbf{18.7} & \textbf{22.3} & &\textbf{86.0}\\
         Topline (\XLSR{}) & & & 19.7 & &100
    \end{tabular}
    }
    \label{tab:2}
\end{table}

\begin{table}[t]
    \centering
    \caption{Transfer of Pre-Trained \WR{} and \XLSR{} models to the target Arabic Language in the MGB dataset}
    \resizebox{.7\linewidth}{!}{%
    \begin{tabular}{lcccc} \toprule
          &  & \multicolumn{2}{c}{WER [\%]}  \\
          \cmidrule{3-4}
         Method & $|\PP|$ [k] & $\PP$ & MGB / ar \\ \midrule
         Baseline (\WR{}) & & & 41.6 \\
         DUST1 & 12 & 21.0& 32.7 \\
         DUST2 & 26& 21.2& 27.4\\
         DUST3 & 30& 20.8& 25.2\\
         DUST4 & 30& 19.5& 23.1\\
         \textbf{DUST5} & 30& \textbf{18.7}& \textbf{21.2}\\
         \XLSR{} & & & 37.9\\
         \midrule
         Baseline (\XLSR{}) & & & 37.9 \\
         DUST1 & 13& 20.3& 31.1\\
         DUST2 & 29& 20.4& 26.3\\
         DUST3 & 30& 20.1& 24.1\\
         DUST4 & 30& 18.5& 22.5\\
         \textbf{DUST5} & 30& \textbf{18.1}& \textbf{20.8}
    \end{tabular}
    }
    \label{tab:3}
\end{table}
In \textbf{Table~\ref{tab:1}}, we show the cross-lingual transferability of different Pre-Trained \WW{} models on eight target languages. The goal is to analyze how much of the multilingual \XLSR{} topline's performance can be recovered by simply Fine-Tuning the English \WW{} models on ten hours of labeled data in target languages. We Fine-Tune a randomly initialized transformer encoder which consists of the same architecture as \WB{} on ten hours of labeled data in each language to use as a baseline. We perform ASR Fine-Tuning of several Pre-Trained English \WW{} on ten hours of labeled data in target languages and compare their ASR performance against the Fine-Tuned \XLSR{} model topline.  We make the following conclusions: 1) \textbf{Pre-Training Matters}: ASR Fine-Tuning of Pre-Trained English \WW{} models lead to significant improvements in WERs on target languages over the baseline. Through the simple \PT{} $\rightarrow$ \FT{} process, we are able to recover on average 79\% to 86\% of the WER and 88\% to 93\% of the CER of the  \XLSR{} topline. 2) \textbf{Big SSL models provide better transfer}: By Fine-Tuning \WL{} that is Pre-Trained on the LS960 dataset, we are able to recover on average 83\% of the topline WER compared to 79\% achieved by Fine-Tuning \WB{} that is also Pre-Trained on LS960. 3) \textbf{Pre-Training dataset size matters upto a point}: Fine-Tuned \WL{} that is Pre-Trained on LL60k recovers on average 86\% of the topline WER compared to 84\% recovered by Fine-Tuning \WL{} that is Pre-Trained on LS960. But the gap in average Word Error Rate Recovery (WERR) between \WR{} that is Pre-Trained on the combined CV, SWBD and LL60k datasets, and \WL{} that is Pre-Trained only on LL60k is less than one percentage point (pp). 4) \textbf{ASR Fine-Tuning of SSL models on source language hurts transfer}: The average WERR on target languages of \WLS{} model which is Pre-Trained on LL60k followed by its ASR Fine-Tuning on labeled LS960 is worse than directly Fine-Tuning the Pre-Trained \WW{} models on the target languages. The WERR for \WLS{} is about 8pp worse than \WR{} that is directly Fine-Tuned on target languages. 5) \textbf{About the out-of-domain Arabic Target Language}: We see that on the seven in-domain languages (MLS/x, where x is the target language) \XLSR{} achieves an average WER of 16.5\% compared to 29.8\% achieved by the ASR Fine-Tuning of \WR{}, the best of the English \WW{} models, giving a performance gap of about 14pp between the two. However, on the out-of-domain Arabic target language (MGB/ar), the gap is less than 4pp. 
Next, we perform cross-lingual adaptation of Pre-Trained \WW{} models using DUST. We choose French and Arabic as the target languages for transfer learning and \WR{} and \XLSR{} as the target models for adaptation. 

In \textbf{Table~\ref{tab:2}}, we use DUST to perform cross-lingual adaptation of Pre-Trained \WR{} to French (MLS/fr). DUST proceeds as follows: 1) First, we perform the ASR Fine-Tuning of the initial \WR{} ($f_{\phi_0, p}$) model using the standard nine hours labeled split provided by MLS/fr dataset to get the first-generation teacher $f_{\phi_1, p}$ (\S \ref{sec:transfer_algo}). 2) Second, $f_{\phi_1, p}$ is used to generate pseudo-labels on the random 100 hours unlabeled split from MLS/fr, which amounts to about 30k utterances, using the pseudo-label generation process explained in \S \ref{sec:transfer_algo} to give a set $\PP$ of pseudo-parallel data. We use 0.2 as the value of the DUST filtering threshold $\tau$. We choose $\tau$ blindly without tuning it on a labeled validation set. 3) Lastly, we Fine-Tune \WR{} (student), $f_{\phi_0, p}$, on the combined labeled and pseudo-labeled data $\PP$ to get $f_{\phi_2, p}$, which is used as the teacher for the next iteration of DUST. We perform a total of five DUST iterations. The final student model $f_{\phi_5, p}$ achieves a WER of 22.3\% which is 16pp lower than the WER of 38.3\% achieved by the first generation teacher model $f_{\phi_1, p}$. Furthermore, $f_{\phi_5, p}$ is able to recover 86\% of the \XLSR{} topline's WER. 
Additionally, we make the following observations: 1) Unsurprisingly, the size of the filtered pseudo-label set $\PP$ (denoted as $|\PP|$ in Table \ref{tab:2}) is larger in later DUST iterations due to the continual improvement in the quality of the student (see WER [\%] in Table \ref{tab:2}), which leads to an improved teacher for subsequent DUST iterations; an improved teacher leads to cleaner pseudo-labels and hence less rejected unlabeled data points during the pseudo-label filtering process. 2) Also, in the later DUST iterations the quality of the pseudo-labels improve, which is implied by the lower WER on pseudo-label set $\PP$ during the later iterations. Next, we consider Arabic (MGB/ar) as the target language for transfer learning, a more challenging transfer learning scenario.

In \textbf{Table~\ref{tab:3}}, we perform adaptation of \WR{} and \XLSR{} to the MGB/ar dataset. Here, the results are achieved by following the same adaptation process detailed above for experiments in Table~\ref{tab:2}. After five DUST iterations, we achieve the final WER of 20.8\% when starting with a Fine-Tuned \XLSR{} model as the first generation teacher $f_{\phi_1, p}$. This result is about 17pp better than the WER of 37.4\% with $f_{\phi_1, p}$. Similar improvements are achieved when using the Fine-Tuned $\tt w2v\_rob$ as $f_{\phi_1, p}$ for DUST iterations.


\section{Conclusions}
We conclude by summarizing the key findings of the paper. We show (Table~\ref{tab:1}) that the monolingual Pre-Trained \WW{} models transfer well across multiple languages. In particular, we show that by performing ASR Fine-Tuning of $\tt wav2vec2\_robust$ on ten hours of labeled data in a target language we are able to recover on average 86\% of the performance of the topline multilingual \XLSR{} model that is Pre-Trained on 53 languages and Fine-Tuned on the same amount of labeled target language data. This finding concurs with similar findings of \cite{rivire2020unsupervised} on cross-lingual transfer of monolingual Pre-Trained SSL models to different target languages for the task of phoneme recognition. Our work goes a step further and proposes a simple yet effective cross-lingual transfer learning algorithm (\S \ref{sec:transfer_algo}) for adaptation of monolingual \WW{} models via Dropout Uncertainty-Driven Self-Training (DUST) by leveraging hundred hours of unlabeled speech data from the target language. We show (Table~\ref{tab:2}) that DUST improves over the baseline model that is Fine-Tuned only on labeled target language data, and is able to recover 86\% of the WER of the topline \XLSR{} model when adapting to French. We show similar results (Table~\ref{tab:3}) when considering Arabic as the target language. Future work should explore combining our method with the adapter framework for cross-lingual transfer learning \cite{pfeiffer2020madx, kannan2019largescale, kessler2021continualwav2vec2, hou2021exploiting}.


\vfill\pagebreak

\bibliographystyle{IEEEtran}
\bibliography{refs}

\end{document}